\documentclass[runningheads]{llncs}
\usepackage{wrapfig}
\usepackage{makeidx}
\usepackage{graphicx}
\usepackage{booktabs}
\usepackage{amsfonts,amssymb}
\usepackage{xcolor}
\usepackage{bbding}
\usepackage{cite}
\usepackage{amsmath}
\usepackage{multirow}
\usepackage{floatrow}
\usepackage[misc]{ifsym}
\usepackage{subfigure}
\newfloatcommand{capbtabbox}{table}[][\FBwidth]
\begin{document}
\title{Doctor Imitator: TW-Imitative Bone Age Assessment Using Hand Radiographs\\ {\small (Original Title: \textit{Doctor Imitator: A Graph-based Bone Age Assessment Framework Using Hand Radiographs})}}
\titlerunning{Doctor Imitator for Bone Age Assessment}  

\renewcommand{\thefootnote}{\fnsymbol{footnote}}
\setcounter{footnote}{-1}

\author{
Jintai Chen\inst{1,2}\footnotemark[1]  \and 
Bohan Yu\inst{1,2}\footnotemark[1] \and  
Biwen Lei\inst{1,2}\footnotemark[1] \and  
Ruiwei Feng\inst{1,2} \and    
Danny Z. Chen\inst{3} \and    
Jian Wu \inst{1,2} \Letter    
}
\authorrunning{Jintai Chen, et al.} 
%
\tocauthor{Jintai Chen, Bohan Yu, et al.}

\footnotetext[1]{These authors contributed equally to this work.}
\institute{
College of Computer Science and Technology, Zhejiang University, Hangzhou, China\\ \email{wujian2000@zju.edu.cn}\\
\and
Real Doctor AI Research Centre, Zhejiang University, Hangzhou, China\\
\and
Department of Computer Science and Engineering, University of Notre Dame, Notre Dame, IN 46556, USA
}
\maketitle      
\begin{abstract}
Bone age assessment is challenging in clinical practice due to the complicated bone age assessment process. Current automatic bone age assessment methods were designed with rare consideration of the diagnostic logistics and thus may yield certain uninterpretable hidden states and outputs. Consequently, doctors can find it hard to cooperate with such models harmoniously because it is difficult to check the correctness of the model predictions. In this work, we propose a new graph-based deep learning framework for bone age assessment with hand radiographs, called Doctor Imitator (DI). The architecture of DI is designed to learn the diagnostic logistics of doctors using the scoring methods (e.g., the Tanner-Whitehouse method) for bone age assessment. Specifically, the convolutions of DI capture the local features of the regions of interest (ROIs) on hand radiographs and predict the ROI scores by our proposed Anatomy-based Group Convolution, summing up for bone age prediction. Besides, we develop a novel Dual Graph-based Attention module to compute patient-specific attention for ROI features and context attention for ROI scores. As far as we know, DI is the first automatic bone age assessment framework following the scoring methods without fully supervised hand radiographs. Experiments on hand radiographs with only bone age supervision verify that DI can achieve excellent performance with sparse parameters and provide more interpretability.\\
\end{abstract}
\section{Introduction}\label{sec:introduction}
\indent Bone age differs from the chronological age and often varies with the gender and ethnicity~\cite{boneagereview}. Bone age assessment (BAA) is typically used to estimate the skeletal maturity and diagnose the growth problem of children. There are two widely employed methods for BAA using hand radiographs in clinical practice: the Greulich-Pyle (GP) method~\cite{todd1950radiographic} and the scoring methods (e.g., the Tanner-Whitehouse method~\cite{tanner1975prediction}). In the GP method, bone age is estimated by referring the entire hand radiographs to the atlas. In the scoring methods, doctors analyze and score the region of interests (e.g., the joints) of the hands individually, and use the weighted sum of these scores to estimate the bone age. It was verified that the scoring methods were more accurate and reliable than the GP method~\cite{boneagereview}.\\
\indent Recently, various automatic BAA methods were proposed~\cite{bonet,superpixel,zhang2019deep,ARCNN}. Similar to the processing of the GP method, most of these models predicted bone ages by capturing the features of the entire hand, which benefited from the capability of deep learning models. Larson et al.~\cite{larson2018performance} built a classifier to predict bone ages, using ResNet50~\cite{he2016deep} as the backbone. Iglovikov et al.~\cite{iglovikov2018paediatric} trained several end-to-end regression models and predicted bone ages using the model ensemble strategy. In~\cite{van2018automated}, the Gaussian process regression was employed to increase the sensibility to the hand pose. Wang et al.~\cite{wang2018bone} followed the structure of Faster-RCNN to predict bone ages. Besides, attention methods were utilized in~\cite{superpixel,ARCNN} to highlight the essential parts of the hands. A relation computing module was introduced in~\cite{PRSNet} to better deal with the important parts of the hands. A new annotation for the Central Positions of Anatomical ROIs (CPAR) to the Radiological Society of North America BAA (RSNA-BAA) dataset was published~\cite{bonet}, which provided the ground truth central positions of the region of interests (ROIs). Using the CPAR ground truth, BoNet~\cite{bonet} promoted the performance of the challenge champion methods. Although these methods could obtain good performance in BAA, they used features of the entire hands and incurred poor interpretability. Following the scoring methods, TW-AI~\cite{yitu} predicted bone ages based on the scores of ROIs, and thus promoted interpretability. However, training TW-AI required full supervision of the ground truth ROI scores, which was uneconomical and hard to popularize. On the other hand, the ROI-based classification method was proved to be feasible in \cite{kazi2017automatic}.\\
\indent In this paper, we propose a novel deep learning framework, called Doctor Imitator (DI), for predicting ROI scores and bone ages using hand radiographs with only bone age supervision. DI is designed by imitating the diagnostic logistics of doctors and the processing of the scoring methods, and obtains excellent and interpretable results with extremely low model complexity. Specifically, we propose an Anatomy-based Group Convolution (AG-Conv) to predict ROI scores using the local features of ROIs and sum up the ROI scores for the bone age prediction. In clinical practice, an experienced doctor may assign the ROI scores with the consideration of some patient-specific characteristics of the bones. Motivated by this, we develop a novel Dual Graph-based Attention Module (DGAM) to assist ROI score prediction, which consists of two new graph-based convolution (GConv) blocks. These two GConv blocks compute a patient-specific attention map for the ROI features and a context attention map for ROI scores, respectively. Different from the previous graph-based convolution (GConv) methods, our new GConv constructs two graphs on one radiograph and updates the features of nodes according to these two graphs simultaneously. Experiments on the RSNA-BAA dataset and a private dataset verify that our DI framework achieves good performance on bone age prediction and ROI score prediction with only bone age supervision.\\
\indent There are three main contributions in this work. \textbf{(A)} We propose a novel deep learning model to predict bone ages and ROI scores, following the diagnostic logistics of doctors and the scoring methods. \textbf{(B)} We introduce a novel dual graph-based attention module to compute patient-specific attention and context attention, updating the node features with two graphs simultaneously. \textbf{(C)} Experiments show that our DI framework can predict ROI scores with only bone age supervision, and thus increase the interpretability of the model.\\
\begin{figure}[tb]
    \centering
    \includegraphics[width=\textwidth]{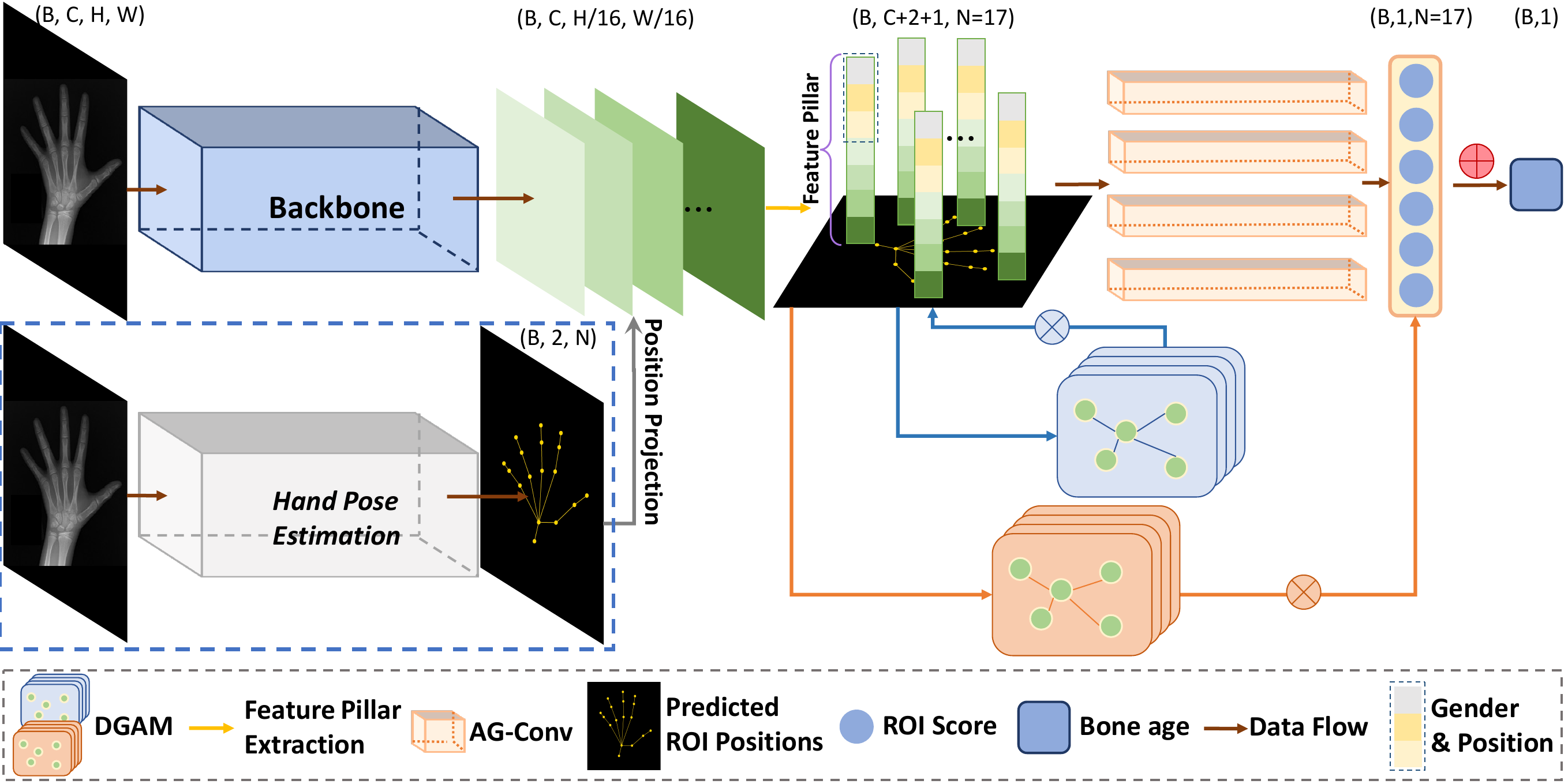}
    \caption{Illustrating our two-stage Doctor Imitator framework. The first stage (\textit{Hand Pose Estimation}), framed in the blue box, predicts the central positions of ROIs. In the second stage, the local features of ROIs are extracted to predict ROI scores, which are summed up for bone age prediction. A Dual Graph-based Attention Module (DGAM) computes the attention maps to help ROI score prediction. The feature sizes are marked above the feature maps.}
    \label{fig:framework}
\end{figure}
\section{Doctor Imitator Architecture}
\indent When estimating bone ages using the scoring methods, doctors often analyze the local characteristics of ROIs (e.g., some joints), and assign a score to every ROI according to the criteria. When assigning scores, experienced doctors also consider some common characteristics of the patient's bones. Finally, the ROI scores are summed up for bone age prediction. Imitating this process, we design a two-stage Doctor Imitator (DI) framework for bone age prediction, as illustrated in Fig.~\ref{fig:framework}. In the first stage, a {\it Hand Pose Estimation} model is trained to predict the central positions of ROIs. Since the first stage is not the focus of this work, we use just the {\it Hand Pose Estimation} model of BoNet~\cite{bonet} by re-implementing it. In the second stage, we train a model to predict ROI scores and bone ages. We take the modified SSN~\cite{SSN} as the backbone, called Improved SSN (ImSSN), to extract the features of hand radiographs. Then we extract the feature pillars (the local features) of the predicted ROIs and perform an Anatomy-based Group Convolution (AG-Conv) module on the ROI feature pillars to predict the ROI scores. During the score prediction, a novel Dual Graph-based Attention Module (DGAM) is utilized to compute patient-specific attention for ROI features and context attention for ROI scores. Finally, we sum up the weighted ROI scores (the weights are provided by context attention) as the predicted bone age. We train the second stage of DI using $L_1$ loss. In our DI framework, there are 17 ROIs (see Fig.~\ref{fig:Anatomy}), since we train the first stage ({\it Hand Pose Estimation}) using the CPAR ground truth data~\cite{bonet}.\\
\indent In what follows, we describe the backbone and feature pillar extraction in Sec.~\ref{sec:featurepillar}, the AG-Conv module for ROI score prediction in Sec.~\ref{sec:AGConv}, and the DGAM for attention computing in Sec.~\ref{sec:dgam}.
\subsection{Backbone and Feature Pillar Extraction}
\label{sec:featurepillar}
\indent In clinical practice, doctors assign ROI scores based on ROI local features. We use SSN~\cite{SSN} with some modification as the backbone, called improved SSN (ImSSN), which is a network similar to U-net and extracts multi-scaled features. We add a $3 \times 3$ average pooling layer on the top-most layer of SSN, and preserve only the $8\times$ and $16\times$ down-sampled branches and output a $16\times$ down-sampled feature map.\\
\indent We use ImSSN to process a hand radiograph and obtain a feature map $F_m \in \mathbb{R}^{C \times H/16 \times W/16}$, where $(H, W)$ is the size of an input hand radiograph, and $C$ is the number of feature channels. To obtain the local features of the ROIs, we project the predicted ROI central positions onto the feature map $F_m$, and take the features on the projected positions as the local features of ROIs. This position projection can be formulated by:
\begin{equation}\label{eq:position}
    (i,j) = (\lfloor\frac{I}{s}\rfloor, \lfloor\frac{J}{s}\rfloor)
\end{equation}
where $(I, J)$ indicates the original central position of an ROI, $(i, j)$ is the corresponding position on the feature map, and $s$ is the down-sampling rate of ImSSN ($s=16$ in this work). As a convolution captures features by fusing the features around, the feature pillars on the projected positions can represent the features of different ROIs. We extract the feature pillars as:
\begin{equation}
    F_p = F_m[:, i, j]
\end{equation}
where $F_p \in \mathbb{R}^{C \times 1 \times 1}$ is the feature pillar of an ROI. Besides, since the gender and ROI central positions are helpful to ROI score prediction, we add a binary value representing the gender and the original ROI central position to the corresponding feature pillar, thus $F_p \in \mathbb{R}^{(C+2+1) \times 1 \times 1}$, as illustrated in Fig.~\ref{fig:framework}.\\
\subsection{Anatomy-based Group Convolution Module}
\label{sec:AGConv}
\begin{figure}[tb]
    \centering
    \includegraphics[width=6.3cm]{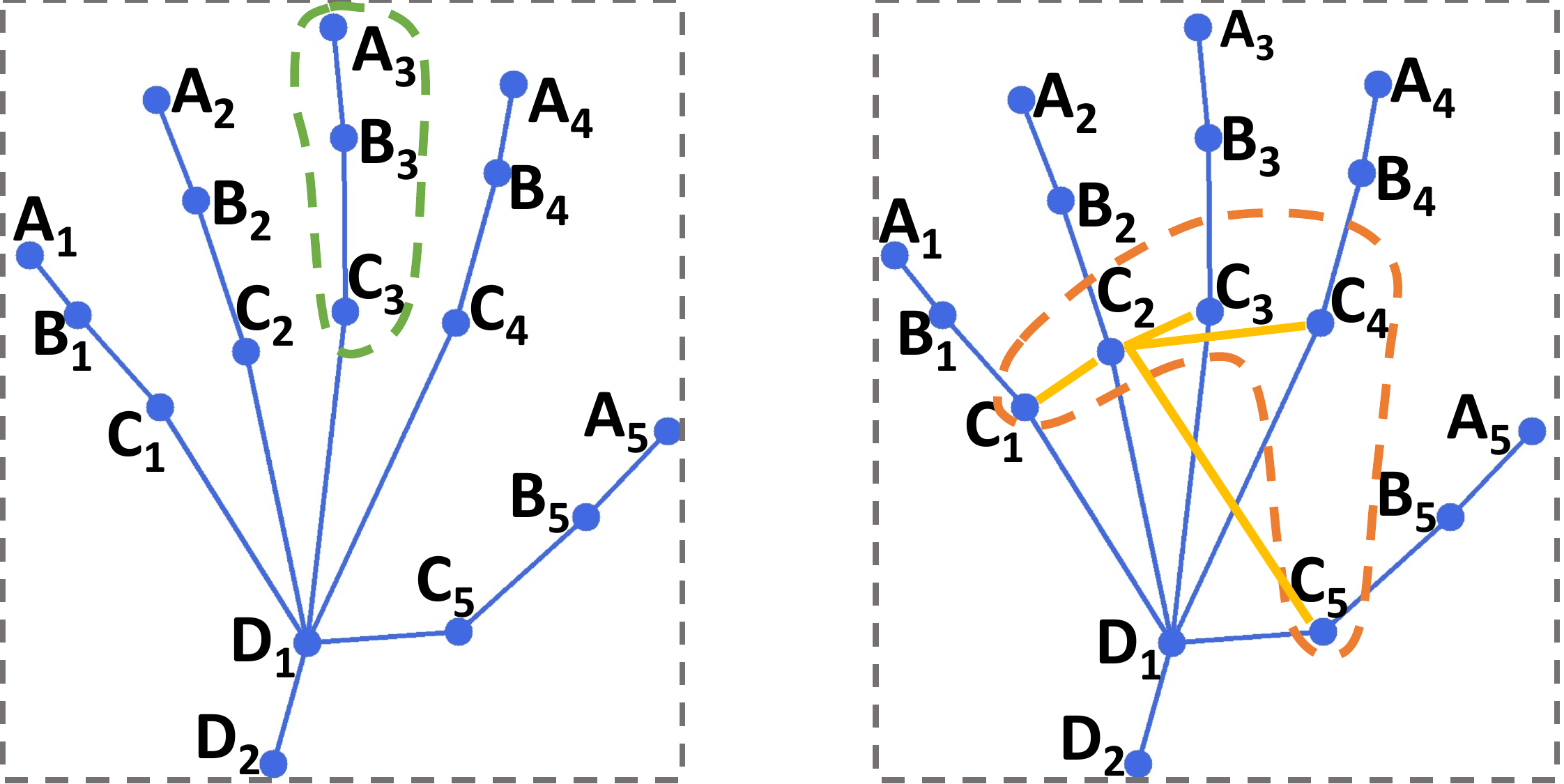}
    \caption{Illustrating the ROI central positions and the receptive fields on different graphs (see Sec.~\ref{sec:dgam}). The receptive field of $\text{B}_3$ on graph $\mathcal{G}_1$ is in the green curve region shown in the left sub-figure. The receptive field of $\text{C}_2$ on graph $\mathcal{G}_2$ is shown in the red curve region in the right sub-figure.}\label{fig:Anatomy}
\end{figure}

\indent In the scoring methods, the ROIs with similar anatomy usually use similar scoring criteria. In Fig.~\ref{fig:Anatomy}, the ROIs marked with the same letters (e.g, $\text{A}_1, \text{A}_2, \text{A}_3, \text{A}_4, \text{A}_5$) are anatomically similar, and the ROIs can be divided into four \textit{anatomy groups}: A, B, C, and D. Following this, we propose an Anatomy-based Group Convolution (AG-Conv) module with four convolution blocks corresponding to the four \textit{anatomy groups}. For the ROIs in an \textit{anatomy group}, one convolution block is used to process the feature pillars and predict the scores. These convolution blocks can be implemented by one-by-one convolutions with batch normalization and ReLU activation, and thus the AG-Conv module shall be very light.\\
\subsection{Dual Graph-based Attention Module}
\label{sec:dgam}
When assigning scores to ROIs, an experienced doctor not only focuses on the local characteristics of bones but also pays attention to some patient-specific characteristics. Besides, the ROI scores are weighted and summed up for bone age prediction in the scoring methods. Following this process, we propose a Dual Graph-based Attention Module (DGAM) with two graph-based Convolution blocks to compute patient-specific attention maps for ROI feature pillars and a context attention map for ROI scores, which are different from the self-attention style in the known graph-based attention methods (e.g., GAT~\cite{GAT}).
\subsection{Graph Construction}
Different from the previous graph-based methods, we construct two undirected graphs $\mathcal{G}_1=(\mathcal{V},\mathcal{E}_1)$ and $\mathcal{G}_2=(\mathcal{V}, \mathcal{E}_2)$ on one hand radiograph with $N$ ROIs. The node set $\mathcal{V}=\{v_i \in \mathbb{R}^{f}\}$ includes all the ROIs presented with the corresponding feature pillars, for $i=1,2,\ldots,N$ and the feature dimension\footnote{For simplicity, the last two dimensions ($1 \times 1$) of the feature pillar $F_p \in \mathbb{R}^{(C+3) \times 1 \times 1}$ are discarded.} $f=C+3$. Graphs $\mathcal{G}_1, \mathcal{G}_2$ have the same node set $\mathcal{V}$, but their edge sets $\mathcal{E}_1$ and $\mathcal{E}_2$ are different. The edge set $\mathcal{E}_1$ contains the natural connections of joints (ROIs), and $\mathcal{E}_2$ contains the full connections among the ROIs in the same {\it anatomy group}. For example, in Fig.~\ref{fig:Anatomy}, the nodes connected with $B_3$ are $A_3$ and $C_3$ in graph $\mathcal{G}_1$, while in graph $\mathcal{G}_2$, nodes $\text{B}_1, \text{B}_2, \text{B}_3, \text{B}_4, \text{B}_5$ (in the same {\it anatomy group}) are connected to one another. Besides, self-connections are also available to all the nodes in both $\mathcal{G}_1$ and $\mathcal{G}_2$. Since the relation among the ROIs in an {\it anatomy group} and the relation among the naturally connected ROIs (joints) are different, constructing two graphs is helpful to model these relations.
\subsection{GConv Blocks}
DGAM consists of two GConv blocks (see Fig.~\ref{fig:framework}), one for patient-specific attention computing, called Patient-specific Attention Block (PAB), and the other for context attention computing, called Context Attention Block (CAB). Both blocks are implemented in a spatial graph-based convolution manner~\cite{wu2019comprehensive} to make the model light. To feed the nodes (ROIs) presented by feature pillars to a GConv, we reformat the ROI feature pillars on a radiograph into a matrix $X_{(N \times f)}$, where $f$ is the feature dimension and $N$ is the number of ROIs (nodes). Given the graphs $\mathcal{G}_1, \mathcal{G}_2$, the GConv operation can be defined by:
\begin{equation}\label{eq:gconv}
    X_{(N \times f_{i+1})} = \frac{1}{2} \sum_{j=\{1,2\}} L_{(N \times N)}^{(j)} X_{(N\times f_i)} W^{(j)}_{(f_i \times f_{i+1})}
\end{equation}
where $j$ indexes the graphs $\mathcal{G}_j$ $(j\in \{1,2\})$. $W^{(j)}$ is a learnable weight matrix for node feature updating and can be implemented by a one-dimensional convolution. The subscripts of the matrices indicate the matrix sizes. $f_i, f_{i+1}$ indicate the feature dimensions before and after feature updating, respectively. $L^{(j)}=[D^{(j)}]^{-\frac{1}{2}} (A^{(j)}+I) [D^{(j)}]^{\frac{1}{2}}$ is the normalized Signless Laplacian matrix on $\mathcal{G}_j$, and $A^{(j)}, D^{(j)}$ are the first order adjacency matrix and the degree matrix, respectively. By Eq.~(\ref{eq:gconv}), the node features are updated by aggregating the features from the neighboring nodes based on the two graphs. Both PAB and CAB are implemented by sequentially stacking the GConvs (as in Eq.~(\ref{eq:gconv})), with different output channel dimensions. Specifically, PAB outputs a feature map in the same size as the input node features as $Att^X \in \mathbb{R}^{N \times f}$, and CAB outputs a feature map $Att^S \in \mathbb{R}^{N \times 1}$.\\
\subsection{Attention Computing}
The patient-specific attention presents the importance of the feature channels. Thus, we apply the node-wise average to compute the patient-specific attention map $\overline{Att}^X_{(N \times f)}$ by:
\begin{equation}
    \overline{att}^X = \frac{1}{N} \sum^N_{n=1} att^X_{n}
\end{equation}
where $[att^X_1, att^X_2, \ldots, att^X_N]=Att^X_{(N \times f)}$ and $\overline{Att}^X_{(N \times f)}=[\overline{att}^X, \overline{att}^X, \ldots, \overline{att}^X]$. $att^X_{n}$ and $\overline{att}^X$ are feature vectors of size $f$. In the scoring methods, the weights of ROIs are the same for every patient of the same gender. Hence, in training we use the exponential moving average (EMA) to compute the general context attention map conditioned by the gender, with the updating parameter $\theta=0.01$, as specified in Eq.~(\ref{eq:average}); In testing, we used the attention maps learned in training stage.
\begin{equation}\label{eq:average}
    \overline{Att}^S_{g} \leftarrow (1-\theta) \overline{Att}^S_{g} + \theta Att^S_{g \mid B}
\end{equation}
where $\overline{Att}^S_{g}$ indicates the general context attention map of the gender $g$ while $Att^S_{g\mid B}$ indicates the average of the predicted context attention of gender $g$ in the batch $B$. $\overline{Att}^S_{g}$ is initialized as $Att^S_{g\mid B=1}$. Then the attention maps are applied to the feature pillars $X_{(N \times f)}$ and the ROI scores $S_{(N \times 1)}$ of one radiograph by:
\begin{equation}
\left\{\begin{aligned}
& X^*_{(N \times f)} = \overline{Att}^X_{(N \times f)} \odot X_{(N \times f)}\\
& S^*_{(N \times 1)} = \overline{Att}^S_{(N \times 1)} \odot S_{(N \times 1)}
\end{aligned}
\right.
\end{equation}
where $\odot$ denotes the Hadamard production, and $X^*_{(N \times f)}$ and $S^*_{(N \times 1)}$ are the feature pillars and ROI scores after attention.\\
\section{Experiments}
\subsection{Dataset}
We evaluate Doctor Imitator (DI) on the RSNA-BAA dataset \cite{bonet}, which contains 12,611 hand radiographs in the training set, 1,425 radiographs in the validation set, and 200 radiographs in the test set. Before processing by DI, all the radiographs are resized to $512 \times 512$. The ground truth bone ages are from 0 to 18 years. Besides, we use the CPAR ground truth in~\cite{bonet} to guide the ROI central position prediction by the {\it Hand Pose Estimation} model~\cite{bonet}. Similar to the previous work, we report the Mean Absolute Difference (MAD) between the predicted bone ages and the corresponding ground truth bone ages.
\subsection{Experimental Setup}
We implement DI by PyTorch 1.3. We train the second stage of DI with 200 epochs. The batch size is 48. The initial learning rate is $10^{-3}$, and is reduced by $10 \times$ after 60 epochs and is reduced by $10 \times$ again after 120 epochs. The optimizer is Adam~\cite{Adam}. We employ random flip, rotation ($-5^{\circ}\sim 5^{\circ}$), and the Gaussian blur Operation for data augmentation.\\
\subsection{Performance and Complexity Comparisons}
We compare DI with the state-of-the-art BoNet~\cite{bonet} on the MAD and model complexity. As shown in Table \ref{table:comparison}, our DI outperforms BoNet on the RSNA training set with various percentage of the training samples available. Comparing to the public implemented version of BoNet~\cite{bonet}, the bone age prediction part of DI is over $12 \times$ smaller in model size than the bone age prediction part of BoNet. Also, evaluated on the number of floating-point multiplication-adds (FLOPs) on our GPU, DI outperforms BoNet by a clear margin. In inference, DI is $\sim 3\times$ faster than BoNet, handling 98.2 bone age radiograph frames per second (fps). Experiments show that DI is very efficient, attains good performance, and has low model complexity.
\begin{table}[tb]
\begin{tabular}{l|cccc|ccc}
\hline
Model & MAD (10\%) & MAD (20\%) & MAD (40\%) & MAD (100\%) & Model Size & FLOPs & fps  \\ \hline
BoNet~\cite{bonet} & 6.78 & 6.18 & 5.31 & 4.37 & 123.2 M  & 17.8  & 39\\
DI (ours)& \textbf{6.42} & \textbf{6.04} & \textbf{5.26} & \textbf{4.30} & \textbf{9.8 M} & \textbf{13.1} & \textbf{98.2}\\
\hline
\end{tabular}
\caption{Comparison with the state-of-the-art BoNet~\cite{bonet} on the performance of the RSNA-BAA test set and the model complexity. The percentages in the parentheses indicate what percentage of the training samples are used in training.}
\label{table:comparison}
\end{table}
\subsection{Doctor Imitating Improvement Study}
\indent DI is designed by imitating the diagnostic logistics of doctors. The experimental results in Table~\ref{table:ablation} verify that imitating doctors obtains improvements. To show the capability of AG-Conv, we compare it with the Random Group Convolution (RG-Conv). In performing RG-Conv, we group the ROIs into four groups randomly and process the ROIs in the groups with the same convolution module. One can see that it is helpful by using the same convolutions on the anatomically similar ROIs. Comparing Exp.4, Exp.5, and Exp.6 in Table~\ref{table:ablation}, it is evident that both of the patient-specific attention and context attention are helpful.\\
\begin{table}[tb]
\begin{tabular}{c|ccccc|c}
\hline
Exp. & ImSSN & AG-Conv & RG-Conv & PA & CA & MAD \\ \hline
1 & \checkmark  &   &    &      &     & 5.38  \\
2 & \checkmark  & \checkmark    &   &      &   &  4.86  \\
3 & \checkmark  &   & \checkmark      &    &   &    4.99 \\
4 & \checkmark  &  \checkmark &    & \checkmark &   &  4.69   \\
5 & \checkmark  &   \checkmark&  &   & \checkmark   &  4.73 \\
6 & \checkmark  &   \checkmark &  &  \checkmark  & \checkmark &  4.48\\ \hline
\end{tabular}
\caption{Ablation study on the proposed modules of DI. ``PA'' denotes the patient-specific attention and ``CA'' denotes the context attention.}
\label{table:ablation}
\end{table}
\begin{figure}[tb]
    \centering
    \includegraphics[width=0.85\textwidth]{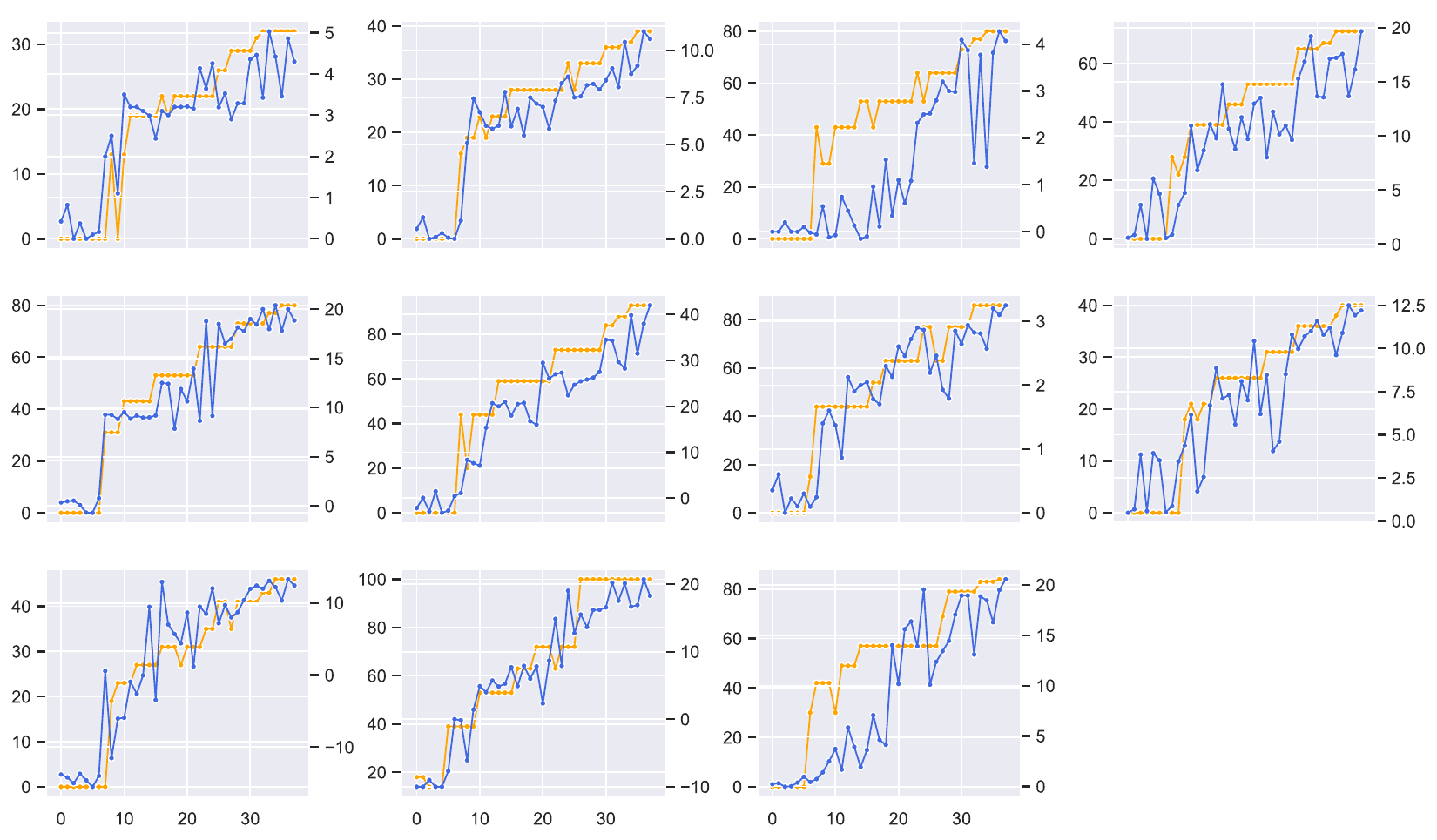}
    \caption{Illustrating the ROI scores (blue points) and the ground truth ROI scores (yellow points). The $x$-axis indexes the patients, while the left $y$-axis shows the ground truth ROI scores and the right $y$-axis shows the predicted ROI scores. The curves are to better show the consistence of the ground truth scores and predicted scores.}
    \label{fig:inter}
\end{figure}
\subsection{Interpretability and Standard ROI Score Prediction}
To evaluate the interpretability of DI, we test DI on a private dataset including 38 hand radiographs with 11 ROI scores following the Chinese-CHN method. DI is first trained on RSNA-BAA dataset and then fine-turned on the private dataset. As shown in Fig.~\ref{fig:inter}, one can see that the predicted ROI score curves highly coincide with the ground truth ROI score curves. Notably, the ROI scores were not used in training, and the results imply the generalization of our DI approach. In practice, the predicted ROI scores can be translated to the standard scores of the scoring methods by a mapping function (similar to the inverse mapping from ROI scores to bone age in the scoring methods). To obtain the standard ROI scores, we simply utilize 11 linear mapping functions to fit these 11 kinds of ROIs respectively, translating the predicted ROI scores to ROI scores, as:
\begin{equation}
    S_i = w_i \times s_i + b_i
\end{equation}
where $S, s$ are the standard ROI scores and predicted scores, respectively, $w$ and $b$ are the weights and the biases of the linear functions, and the subscript $i$ indexes the ROIs. The weights, biases, p values of the F-statistics, and results of the adjusted $R^2$ are reported in Table~\ref{tab:test}. The p values of the F-statistics verify that the mapping functions are statistically significant, and the results of the adjusted $R^2$ show that the ground truth ROI scores (standard ROI scores) are well explained by the predicted ROI scores. In clinical practice, it is recommended to use more complicated fitting functions.
\begin{table}[b]
\begin{tabular}{c|c|c|c|c}
\hline
ROI Index & weight & bias & p values of F-statistic & adjusted $R^2$ \\ \hline
1         & 7.38 & -2.23 & $2.612\times 10^{-15}$    & 0.8228         \\
2         & 3.75 & 0.96 & $\textless{}2\times 10^{-16}$    & 0.9048 \\
3         & 12.8 & 29.30 & $1.312\times 10^{-7}$    & 0.5306         \\
4         & 3.89 & 3.17 & $4.364\times 10^{-14}$   & 0.793          \\
5         & 3.74 & 5.68 & $\textless{}2\times 10^{-16}$  & 0.8882         \\
6         & 2.22 & 12.10& $\textless{}2\times 10^{-16}$   & 0.8845         \\
7         & 4.71 & 25.57 & $2.209\times 10^{-15}$  & 0.8244         \\
8         & 3.26 & 0.88 & $7.21\times 10^{-13}$  & 0.7584         \\
9         & 1.43 & 23.72 & $3.914\times 10^{-15}$    & 0.8188         \\
10        & 2.79 & 45.17& $\textless{}2\times 10^{-16}$   & 0.8807         \\
11        & 2.96 & 24.56 & $4.791\times 10^{-10}$       & 0.6546         \\ \hline
\end{tabular}
\caption{The weights, biases, p values of the F-statistics, and the results of the adjusted $R^2$ of the mapping functions.}
\label{tab:test}
\end{table}
\section{Conclusions}
\indent In this paper, we proposed an automatic bone age assessment model, Doctor Imitator (DI), by imitating the diagnostic logistics of doctors using the scoring methods. An Anatomy-based Group Convolution was proposed to predict the ROI scores by processing the local features of ROIs. Besides, a novel Dual Graph-based Attention Module was introduced to compute the patient-specific attention and context attention for ROI score prediction. As far as we know, DI is the first BAA framework following the processing of the scoring methods with only bone age supervision. Compared with the state-of-the-art, DI achieves good performance with low model complexity and excellent interpretability.\\
\section{Acknowledgements}
The research of Real Doctor AI Research Centre was partially supported by the Zhejiang University Education Foundation under grants No.K18-511120-004, No.K17-511120-017, and No.K17-518051-021, the National Natural Science Foundation of China under grant No.61672453, the National key R\&D program sub project ``large scale cross-modality medical knowledge management'' under grant No.2018AAA0102100, the Zhejiang public welfare technology research project under grant No.LGF20F020013, the National Key R\&D Program Project of ``Software Testing Evaluation Method Research and its Database Development on Artificial Intelligence Medical Information System'' under the Fifth Electronics Research Institute of the Ministry of Industry and Information Technology (No.2019YFC0118802), and The National Key R\&D Program Project of ``Full Life Cycle Detection Platform and Application Demonstration of Medical Artificial Intelligence Product'' under the National Institutes for Food and Drug Control (No.2019YFB1404802), and the Key Laboratory of Medical Neurobiology of Zhejiang Province. The research of D.Z. Chen was partially supported by NSF Grant CCF-1617735.\\
\bibliographystyle{splncs04}
\bibliography{references}
\newpage
\appendix
\section{Some Implementation Details}
\subsection{ImSSN Structure}
\begin{figure}
    \centering
    \includegraphics[width=0.7\textwidth]{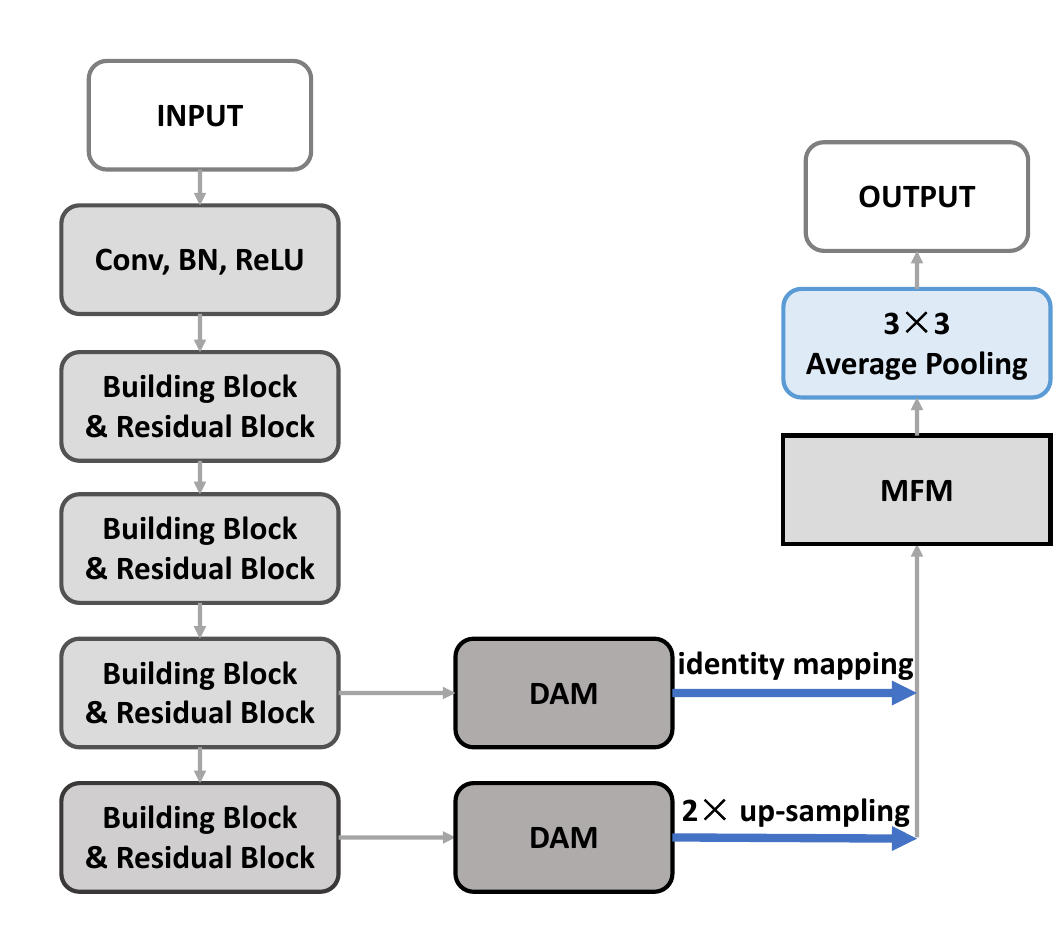}
    \caption{An illustration of ImSSN. The modified parts are marked in blue color.}\label{imssn}
\end{figure}

\indent An implementation version of ImSSN is shown in Fig.~\ref{imssn}. The names of the modules in ImSSN follow those from SSN~\cite{SSN}. 
\subsection{Structure of the Anatomy-based Group Convolution (AG-Conv) Module}
\indent In the AG-Conv module, there are four identical AG-Conv blocks. The detailed structure of an AG-Conv block is provided in Table~\ref{tab:agconv}. The ROI feature pillars with $C+3$ ($C=64$ in this work) feature channels are fed to the AG-Conv blocks, while the output is the ROI scores with one feature channel.
\begin{table}
\centering
\resizebox{\textwidth}{11mm}{
\begin{tabular}{c|l}
\hline
Layer Index & \multicolumn{1}{c}{Layer Structure}                                            \\ \hline
1           & Conv1D (kernel size=1, input channel=64+3, output channel=32, stride=1) + BN + ReLU \\ \hline
2           & Conv1D (kernel size=1, input channel=32, output channel=32, stride=1) + BN + ReLU   \\ \hline
3           & Conv1D (kernel size=1, input channel=32, output channel=64, stride=1) + BN + ReLU   \\ \hline
4           & Conv1D (kernel size=1, input channel=64, output channel=64, stride=1) + BN + ReLU   \\ \hline
5           & Conv1D (kernel size=1, input channel=64, output channel=1, stride=1)            \\ \hline
\end{tabular}}
\caption{The structure of an Anatomy-based Group Convolution block.}\label{tab:agconv}
\end{table}
\subsection{Structure of the Dual Graph-based Attention Module (DGAM)}
\indent In DGAM, the patient-specific attention block (PAB) and context attention block (CAB) are implemented by sequentially stacking model layers. A model layer consists of the proposed graph-based convolution (in Eq.~(3)) and a ReLU activation. The feature sizes are reported in Table~\ref{GCN}.
\begin{table}
\resizebox{\textwidth}{16mm}{
\begin{tabular}{c|c|c|cc}
\hline
\multirow{2}{*}{\begin{tabular}[c]{@{}l@{}}Layer \\ Index\end{tabular}} & \multicolumn{2}{c|}{PAB}                       & \multicolumn{2}{c}{CAB}                                             \\ \cline{2-5} 
                                                                          & Size of input feature & Size of output feature & \multicolumn{1}{c|}{Size of input feature} & Size of output feature \\ \hline
1 & (B, f=64+3, N=17)     & (B, f=64, N=17)        & \multicolumn{1}{c|}{(B, f=64+3, N=17)}     & (B, f=64, N=17)        \\ \hline
2 & (B, f=64, N=17)       & (B, f=64, N=17)        & \multicolumn{1}{c|}{(B, f=64, N=17)}       & (B, f=64, N=17)        \\ \hline
3    & (B, f=64, N=17)       & (B, f=128, N=17)       & \multicolumn{1}{c|}{(B, f=64, N=17)}       & (B, f=128, N=17)       \\ \hline
4  & (B, f=128, N=17)      & (B, f=128, N=17)       & \multicolumn{1}{c|}{(B, f=128, N=17)}      & (B, f=128, N=17)       \\ \hline
5   & (B, f=128, N=17)      & (B, f=256, N=17)       & \multicolumn{1}{c|}{(B, f=128, N=17)}      & (B, f=256, N=17)       \\ \hline
6   & (B, f=256, N=17) & (B, f=64+3 N=17) & \multicolumn{1}{c|}{(B, f=256, N=17)}& (B, f=1, N=17)         \\ \hline
\end{tabular}}
\caption{The feature sizes in the blocks of PAB and CAB. ``B'' indicates the batch size, ``f'' indicates the number of the feature channels, and N is the number of the nodes (ROIs).}
\label{GCN}
\end{table}
\end{document}